\documentclass[10pt,twocolumn]{article}

\usepackage[utf8]{inputenc}
\usepackage[T1]{fontenc}
\usepackage{lmodern}
\usepackage{microtype}

\usepackage[a4paper,margin=1.9cm,columnsep=0.7cm]{geometry}
\usepackage{titlesec}

\usepackage{amsmath}
\usepackage{amssymb}
\usepackage{amsthm}
\usepackage{bm}

\usepackage{booktabs}
\usepackage{array}
\usepackage{graphicx}
\usepackage{float}
\usepackage{caption}
\usepackage{xcolor}
\usepackage{colortbl}
\usepackage[numbers,sort&compress]{natbib}

\usepackage[colorlinks=true,linkcolor=black,citecolor=black,urlcolor=blue!60!black]{hyperref}

\newcommand{\SFTID}{\ensuremath{\mathrm{SFT}_{\mathrm{ID}}}}
\newcommand{\SFTOOD}{\ensuremath{\mathrm{SFT}_{\mathrm{OOD}}}}
\newcommand{\E}{\mathbb{E}}
\newcommand{\KL}{\mathrm{KL}}
\newcommand{\Hop}{\mathcal{H}}

\titleformat{\section}{\normalfont\large\bfseries}{\thesection}{0.6em}{}
\titleformat{\subsection}{\normalfont\normalsize\bfseries}{\thesubsection}{0.5em}{}
\titlespacing*{\section}{0pt}{1.6ex plus 1ex minus .2ex}{1.0ex plus .2ex}
\titlespacing*{\subsection}{0pt}{1.2ex plus 1ex minus .2ex}{0.7ex plus .2ex}

\newcolumntype{C}{>{\centering\arraybackslash}X}

\begin{document}

\twocolumn[
\begin{@twocolumnfalse}
\begin{center}
{\LARGE\bfseries Stage-1 Controls the Entropy Regime, Not the Outcome\par}
\vspace{0.4em}
{\large On-Policy Distillation versus SFT as a Warm-Start for Small-Data VLM RL\par}
\vspace{0.8em}
{\normalsize Jianxiong Shen\par}
\vspace{1.2em}
\end{center}

\begin{center}{\large\bfseries Abstract}\end{center}
\noindent
Two-stage post-training --- a Stage-1 warm-start (supervised fine-tuning, SFT, or on-policy distillation, OPD) followed by Stage-2 reinforcement learning (RL) --- is increasingly used for vision-language models (VLMs). We ask what Stage-1 actually controls in a small-data study using Qwen2.5-VL-7B with a same-modality 72B VLM teacher for OPD. First, the three warm-starts reach a narrow $53$--$54\%$ band on Geometry3K internal validation, consistent with the narrow range reported by recent specialized methods; this setup provides little evidence that Stage-1 changes the in-domain endpoint. Second, a matched-recipe, early-stopped SFT improves out-of-domain MathVista by $+2.1$ points, reversing the $-9.5$-point drop of an over-trained variant. The clearest difference is the \emph{entropy regime}: OPD enters RL with substantially higher policy entropy than either SFT initialization, and the separation remains visible through the available trajectories. At the in-domain initialization, OPD also has higher answer diversity and pass@16 ($+2.0$ to $+5.2$ points over SFT), although problem-level bootstrap intervals show that the smaller contrast is uncertain. The advantage is absent after RL (endpoint pass@16 values within $1.1$ points) and on MathVista (six models within $1.2$ points). Our contribution is therefore a bounded empirical characterization: Stage-1 is strongly associated with the entropy regime in this setup, but the downstream payoff is small, localized, and not evidence that OPD is a better RL warm-start.
\vspace{1.2em}
\end{@twocolumnfalse}
]

\section{Introduction}

Post-training of vision-language models increasingly follows a two-stage template: a Stage-1 warm-start that injects task structure, then Stage-2 RL (typically GRPO-style) that optimizes a verifiable reward. For the warm-start, two families dominate --- supervised fine-tuning (SFT) on curated traces, and on-policy distillation (OPD), which trains the student on its own rollouts under a teacher's token distribution. The choice is often treated as important and is defended by two recurring arguments: that SFT induces \emph{catastrophic forgetting} of base capabilities, and that one warm-start yields a \emph{better initialization} for RL.

We interrogate these claims in the \textbf{small-data regime} ($\approx 1$--$2$k examples) on Qwen2.5-VL-7B \citep{qwen25vl}. Our OPD uses a \textbf{same-modality} teacher --- a 72B Qwen2.5-VL --- distinguishing it from cross-modal (text-teacher) distillation. We compare three warm-starts followed by closely matched GRPO recipes, evaluated in-domain (Geometry3K) and out-of-domain (MathVista \citep{mathvista}). Our findings are:

\begin{itemize}\setlength{\itemsep}{2pt}
\item \textbf{(F1) The observed forgetting is recipe-dependent.} A matched-recipe, early-stopped SFT improves OOD MathVista by $+2.1$ over base; the $-9.5$ drop appears only in the longer, prompt-mismatched variant.
\item \textbf{(F2) In-domain endpoints occupy a narrow band.} The three warm-starts reach $53$--$54\%$ best internal validation, similar to the range reported by recent specialized methods. This is evidence of limited headroom in this setup, not a universal saturation claim.
\item \textbf{(F3, core) Stage-1 is associated with distinct entropy regimes, with bounded payoff.} OPD trajectories have substantially higher entropy than the two SFT trajectories. The corresponding pass@$k$ advantage appears only at the in-domain initialization and is absent after RL and OOD.
\end{itemize}

The contribution is not a method or a leaderboard number but a \textbf{characterization}: a clean separation of what Stage-1 does and does not control, with the entropy regime identified as the single reliable axis and its consequences honestly bounded.

\section{Setup and Notation}

\noindent\textbf{Policy and warm-start.} Let $x=(\text{image},\text{question})$ and let $\pi_\theta(y\mid x)$ be the autoregressive policy over responses $y=(y_1,\dots,y_L)$, with token conditionals $\pi_\theta(y_t\mid h_t)$ on history $h_t=(y_{<t},x)$ over vocabulary $\mathcal V$. A warm-start maps a base policy to an initialization $\pi_0$ for RL. We report the \emph{policy entropy}, the expected per-token conditional entropy under on-policy rollouts,
\begin{equation*}
\Hop(\pi_\theta)=\E_{x}\,\E_{y\sim\pi_\theta(\cdot\mid x)}\!\Big[\tfrac{1}{L}\textstyle\sum_{t=1}^{L} H\!\big(\pi_\theta(\cdot\mid h_t)\big)\Big],
\end{equation*}
\begin{equation*}
H\!\big(\pi_\theta(\cdot\mid h_t)\big)=-\!\sum_{v\in\mathcal V}\!\pi_\theta(v\mid h_t)\log\pi_\theta(v\mid h_t).
\end{equation*}

\noindent\textbf{Model and teacher.} Student: Qwen2.5-VL-7B-Instruct. Teacher: Qwen2.5-VL-72B (same modality). OPD is reverse-KL, token-level, on-policy distillation over all $2{,}101$ Geometry3K training prompts.

\noindent\textbf{Warm-starts.} We compare three initializations $\pi_0$:
\begin{itemize}\setlength{\itemsep}{2pt}
\item \textbf{OPD} --- on-policy distillation from the 72B same-modality teacher over all $2{,}101$ prompts.
\item $\bm{\SFTID}$ --- in-domain-optimized SFT (with task system prompt); strongest in-domain SFT recipe.
\item $\bm{\SFTOOD}$ --- OOD-optimized SFT (no system prompt, early-stopped at $100$ steps); strongest OOD-retention recipe.
\end{itemize}

\noindent\textbf{Same teacher, unequal difficulty coverage (made explicit).} Both warm-starts derive from the \emph{same} 72B teacher: the SFT traces are that teacher's $n{=}4$ rollouts kept when correct (rejection sampling). This removes any teacher-identity confound. The two do not see the same data, however: rejection sampling retains only the \textbf{807 problems} the teacher solves within $4$ samples ($1{,}083$ correct traces, $1.34$ per problem), whereas OPD trains over \textbf{all $2{,}101$} --- including the \textbf{$1{,}294$ problems the teacher never solves}. SFT thus never sees the hard tail OPD does. We treat this as a known, deliberate setting difference, analyze its bearing on F3 in \S6.4, and mark the data-aligned control as the primary follow-up (\S8).

\noindent\textbf{Stage-2 RL.} The runs use the same intended GRPO recipe: no KL penalty, lr $5\times10^{-7}$, train batch $8$, and group size $n{=}5$. The two SFT-initialized production runs complete $786$ steps. OPD evidence combines a historical $786$-step run with a production-stack replication observed through step $680$; endpoint comparisons therefore match the optimization recipe but not every software-stack detail. For a prompt $x$ with sampled group $\{y^{(i)}\}_{i=1}^n$ and rewards $r^{(i)}$, GRPO uses group-relative advantages $\hat A^{(i)}=\big(r^{(i)}-\mathrm{mean}_j r^{(j)}\big)/\mathrm{std}_j r^{(j)}$.

\noindent\textbf{Benchmarks and grading.} In-domain: Geometry3K ($601$-item test). OOD: MathVista testmini ($1{,}000$). Grading: symbolic match (mathruler) for free-form, normalized letter-match for multiple choice.

\noindent\textbf{Evaluation calibration.} In-domain Geometry3K is reported via \emph{training-time internal validation} (the verl rollout engine, identical across runs); OOD MathVista via \emph{offline} evaluation. We verified that the $\approx5$-point internal$\leftrightarrow$offline Geometry3K gap is \textbf{not} a version/environment artifact: re-running offline under the exact training stack recovers only $0.8$ points (a known vLLM-kernel difference), leaving $\approx5$ points to rollout-vs-offline code paths. MathVista shows no such gap. Each comparison therefore stays within a single pipeline; we never cross pipelines within a comparison.

\noindent\textbf{Exploration metric.} With $n{=}16$ samples at temperature $0.8$ and $c$ correct, we report the unbiased estimator
\begin{equation*}
\text{pass@}k=\E_{x}\!\left[\,1-\frac{\binom{n-c}{k}}{\binom{n}{k}}\,\right],
\end{equation*}
a clean greedy@1 (temperature $0$), and answer diversity $\mathcal D=\E_x\big[\lvert\mathrm{unique}(\{\hat a^{(i)}\})\rvert/n\big]$ over extracted answers. pass@1 self-checks against greedy numbers.

\section{Objective-Level Entropy Intuition}

The objectives provide a useful intuition for the observed entropy difference, but do not imply an ordering for finite-capacity models trained on different state distributions.

\noindent\textbf{SFT: cross-entropy to a point mass.} Given a correct teacher trace $y^\star$, SFT minimizes
\begin{equation*}
\mathcal L_{\mathrm{SFT}}(\theta)=-\sum_{t}\log\pi_\theta\!\big(y^\star_t\mid h^\star_t\big),
\end{equation*}
i.e.\ forward cross-entropy to the Dirac target $\delta_{y^\star_t}$ at each visited state. Its per-state minimizer is $\pi_\theta(\cdot\mid h^\star_t)\!\to\!\delta_{y^\star_t}$, so the conditional entropy collapses, $H(\pi_\theta(\cdot\mid h^\star_t))\to 0$. The SFT stationary policy is deterministic on its training support.

\noindent\textbf{OPD: reverse-KL to the teacher.} With the student sampling $y\sim\pi_\theta$ and teacher $\pi_T$, token-level OPD minimizes
\begin{equation*}
\mathcal L_{\mathrm{OPD}}(\theta)=\E_{y\sim\pi_\theta}\sum_t \KL\!\big(\pi_\theta(\cdot\mid h_t)\,\Vert\,\pi_T(\cdot\mid h_t)\big),
\end{equation*}
($+$ optional PG term). For an unrestricted policy at a fixed state, the unique minimizer is $\pi_\theta(\cdot\mid h_t)=\pi_T(\cdot\mid h_t)$. Unlike SFT on one trace, this target need not be a point mass.

\noindent\textbf{Scope of the argument.} This fixed-state comparison is not an entropy lower bound for the trained student. Reverse-KL is mode-seeking; finite capacity, optimization, and student-induced state visitation can all produce entropy below the teacher's, including collapse \citep{decoupling2026}. The objective only supplies a plausible mechanism: when the teacher is non-degenerate on student-visited states, OPD can preserve soft alternatives that single-trace SFT does not expose. Whether it does so is an empirical question. Our results establish the ordering for these runs, not as a universal property of OPD or of same-modality teachers.

\section{Forgetting as a Regime Artifact (F1)}

Scanning SFT recipes while isolating learning rate, packing, system prompt, and step count gives OOD retention on MathVista testmini (offline; base $=68.60$), shown in Table~\ref{tab:forgetting}.

\begin{table}[t]
\centering
\caption{SFT recipe scan: OOD retention on MathVista (offline; base $=68.60$).}
\label{tab:forgetting}
\scriptsize
\setlength{\tabcolsep}{3pt}
\begin{tabular}{@{}llccc@{}}
\toprule
SFT variant & recipe & step & MathVista & $\Delta$ vs base \\
\midrule
over-trained & no sys.\ prompt & 200 & 59.10 & $\mathbf{-9.5}$ \\
$\SFTOOD$ & same, early stop & 100 & $\mathbf{70.70}$ & $\mathbf{+2.1}$ \\
$\SFTID$ & $+$ system prompt & 100 & 65.40 & $-3.2$ \\
OPD-Stage1 & --- & --- & 68.30 & $-0.3$ \\
\bottomrule
\end{tabular}
\end{table}

The over-trained variant and $\SFTOOD$ share the same base recipe and differ in training duration ($200$ vs $100$ steps); MathVista moves $+11.6$. Together with the system-prompt scan, this shows that the observed retention loss is recipe-dependent. In this setup, early-stopped SFT does not degrade OOD, so the over-trained run alone cannot support a general claim that small-data SFT necessarily forgets.

\section{Narrow In-Domain Endpoint Band (F2)}

Under the matched GRPO recipe, the three warm-starts reach a narrow in-domain band on Geometry3K training-time internal validation (best step), Table~\ref{tab:saturation}. We do not have independent training seeds, so ``statistically indistinguishable'' would be too strong.

\begin{table}[t]
\centering
\caption{In-domain GRPO endpoints (Geometry3K internal validation, best step).}
\label{tab:saturation}
\small
\begin{tabular}{@{}lc@{}}
\toprule
warm-start $\to$ GRPO & best internal val \\
\midrule
OPD $\to$ GRPO & $0.536$ (step 559) \\
$\SFTID \to$ GRPO & $0.538$ (step 340) \\
$\SFTOOD \to$ GRPO & $0.531$ (step 340) \\
\bottomrule
\end{tabular}
\end{table}

These values resemble the narrow band reported by specialized methods on Qwen2.5-VL-7B + Geometry3K --- vanilla GRPO $53.3$, Curr-ReFT $53.2$, and GMPO $54.7$ \citep{currreft2025,gmpo2025}. Because splits, graders, and implementations are not identical, this is a band-level comparison only. The evidence suggests limited headroom for separating warm-starts on this benchmark; it does not establish an absolute task ceiling.

\section{The Entropy Regime and Its Bounded Payoff (F3)}

\subsection{Entropy gap and persistence}

\begin{figure}[t]
\centering
\includegraphics[width=\linewidth]{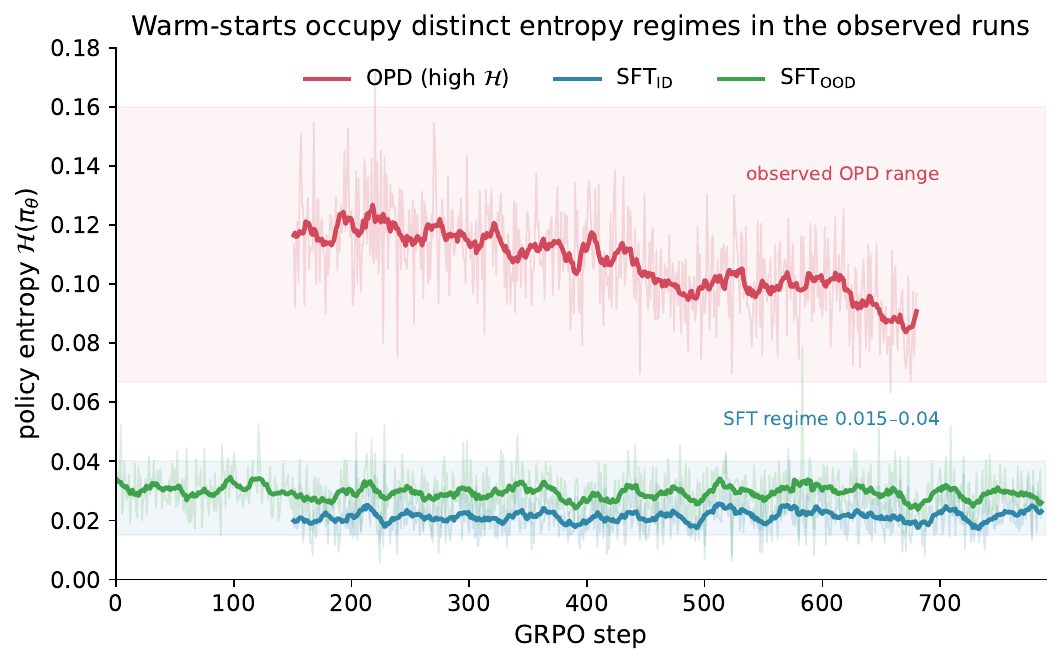}
\caption{Available entropy trajectories under the matched GRPO recipe. The SFT runs complete $786$ steps; the production-stack OPD replication is available through step $680$. OPD declines late in training but remains above the SFT ranges.}
\label{fig:entropy}
\end{figure}

Policy entropy during GRPO confirms the \S3 prediction (Table~\ref{tab:entropy}, Figure~\ref{fig:entropy}).

\begin{table}[t]
\centering
\caption{Observed RL policy-entropy ranges by warm-start.}
\label{tab:entropy}
\small
\begin{tabular}{@{}lc@{}}
\toprule
warm-start & RL policy entropy \\
\midrule
OPD & early: $\mathbf{0.10}$--$\mathbf{0.16}$ \\
 & production run late: $\mathbf{0.067}$--$\mathbf{0.11}$ \\
$\SFTID/\SFTOOD$ & $\mathbf{0.015}$--$\mathbf{0.04}$ \\
\bottomrule
\end{tabular}
\end{table}

OPD starts roughly $5$--$8\times$ above the SFT range. In the production-stack replication it declines to approximately $0.067$--$0.11$ near steps $675$--$680$, still above the contemporaneous SFT range. Similar high-entropy OPD behavior appears in two historical stacks, which supports robustness to implementation details but does not isolate the cause from the data-coverage difference.

\subsection{In-domain exploration}

\begin{figure}[t]
\centering
\includegraphics[width=\linewidth]{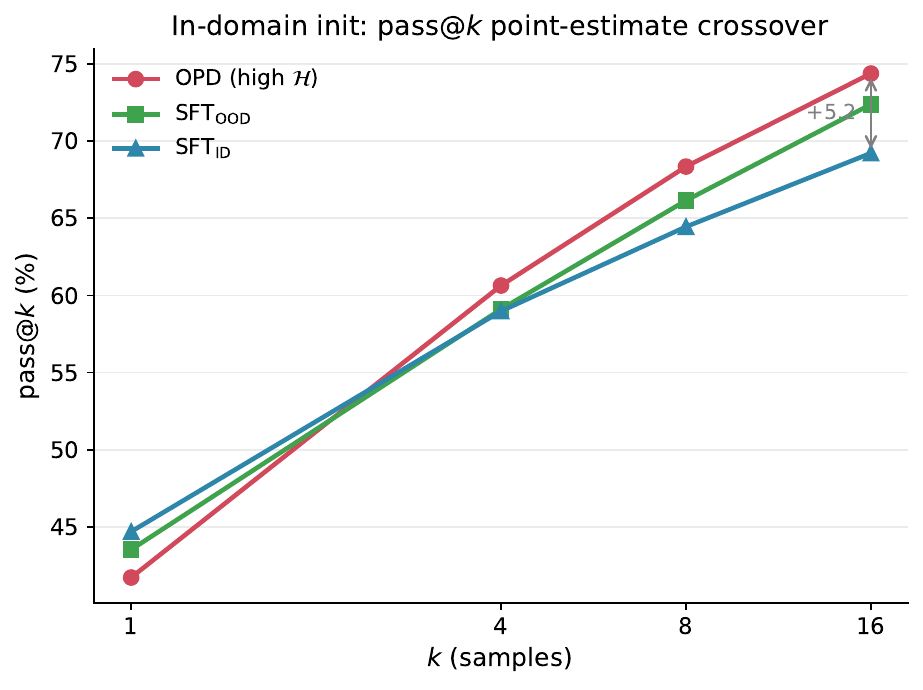}
\caption{pass@$k$ at the in-domain initialization. SFT leads at greedy/pass@1, but the high-entropy OPD warm-start overtakes from pass@4 and widens to $+5.2$ over $\SFTID$ at pass@16.}
\label{fig:crossover}
\end{figure}

We probe the downstream meaning of entropy with pass@$k$ on Geometry3K (offline pipeline). At the \textbf{initialization}, the high-entropy OPD warm-start has a higher exploration ceiling than the low-entropy SFT ones, \emph{even where greedy ties or loses} (Table~\ref{tab:indomain}, Figure~\ref{fig:crossover}).

\begin{table}[t]
\centering
\caption{In-domain pass@$k$ at initialization (Geometry3K, offline).}
\label{tab:indomain}
\small
\setlength{\tabcolsep}{3.5pt}
\begin{tabular}{@{}lcccccc@{}}
\toprule
init & gr@1 & p@1 & p@4 & p@8 & p@16 & div. \\
\midrule
OPD (hi $\Hop$) & 43.9 & 41.7 & 60.6 & 68.3 & $\mathbf{74.4}$ & $\mathbf{28.1}$ \\
$\SFTOOD$ (lo) & 43.6 & 43.5 & 59.1 & 66.1 & 72.4 & 24.2 \\
$\SFTID$ (lo) & $\mathbf{45.9}$ & 44.7 & 59.0 & 64.4 & 69.2 & 21.9 \\
\bottomrule
\end{tabular}
\end{table}

A crossover appears in the point estimates: SFT leads at greedy/pass@1, while OPD overtakes from pass@4 and reaches $+2.0$ (vs $\SFTOOD$) / $+5.2$ (vs $\SFTID$) at pass@16. Paired problem-bootstrap analysis qualifies this result: the OPD--$\SFTID$ pass@16 contrast is supported, while the smaller OPD--$\SFTOOD$ contrast has a confidence interval crossing zero. Policy entropy, answer diversity, and pass@16 point estimates rank identically, but only the larger contrast should be treated as clear evidence.

\subsection{Vanishing under RL and OOD}

\begin{figure}[t]
\centering
\includegraphics[width=\linewidth]{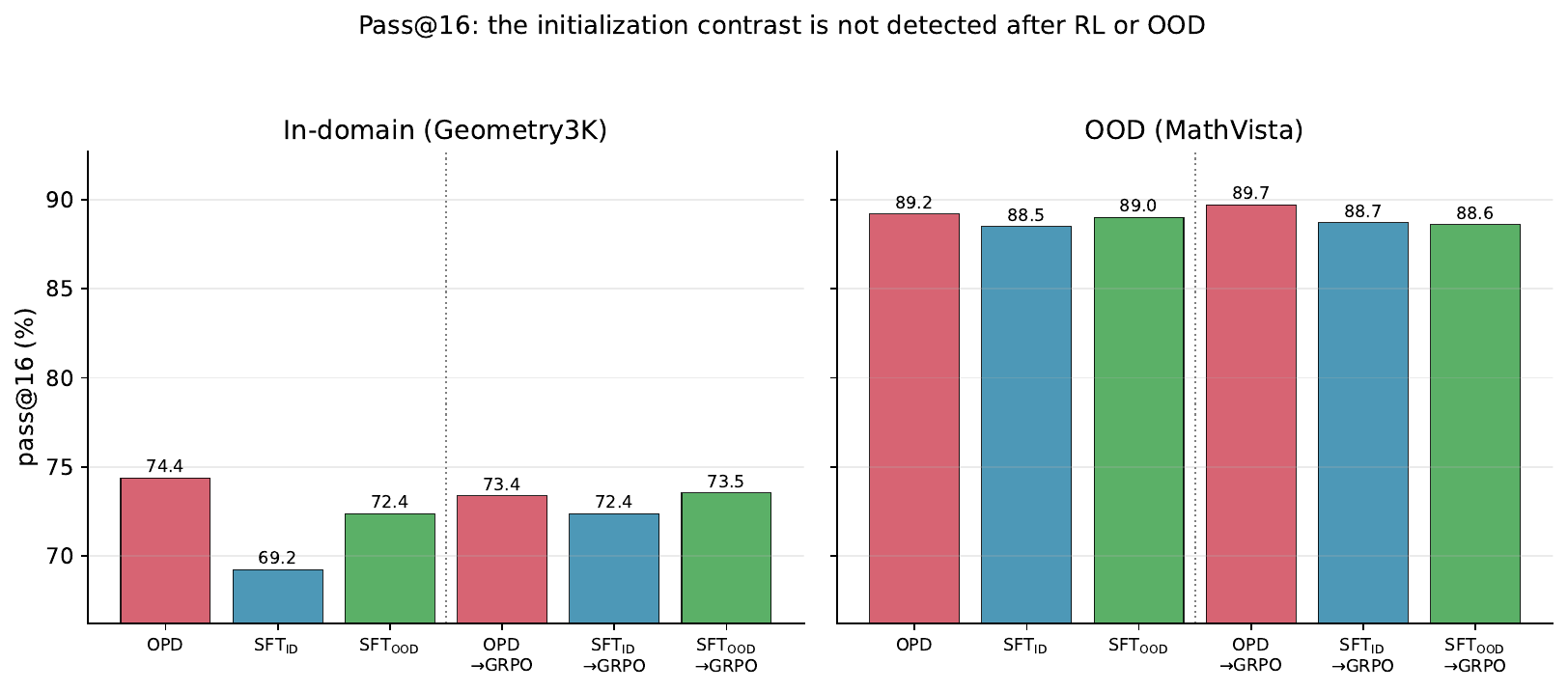}
\caption{pass@16 in-domain (left) vs OOD (right) for all six models. The high-entropy OPD advantage is present only at the in-domain initialization; it is consumed by RL and absent on MathVista.}
\label{fig:ood}
\end{figure}

The advantage is bounded on both testable axes (Figure~\ref{fig:ood}).

\noindent\emph{Absent at the measured RL endpoints.} At the GRPO endpoints (Geo3K), pass@16 point estimates lie within $1.1$: OPD$\to$GRPO $73.4$, $\SFTID\to$GRPO $72.4$, $\SFTOOD\to$GRPO $73.5$. All paired confidence intervals for these endpoint differences cross zero. The current evidence therefore does not show that the initialization advantage survives RL.

\noindent\emph{No detected OOD transfer.} On MathVista, all six point estimates fall within $1.2$ at pass@16 ($88.5$--$89.7$), and all relevant paired confidence intervals cross zero. Diversity lies in a narrow $15.5$--$17.2$ range and the in-domain ordering disappears; see Table~\ref{tab:ood}.

\begin{table}[t]
\centering
\caption{OOD pass@$k$ on MathVista. All six models within $1.2$ at pass@16; entropy ordering disappears.}
\label{tab:ood}
\small
\setlength{\tabcolsep}{3.5pt}
\begin{tabular}{@{}lccccc@{}}
\toprule
model (MathVista) & gr@1 & p@4 & p@8 & p@16 & div. \\
\midrule
OPD init & 68.2 & 82.3 & 86.5 & $\mathbf{89.2}$ & 16.9 \\
$\SFTOOD$ init & $\mathbf{69.0}$ & 82.7 & 86.6 & 89.0 & 16.2 \\
$\SFTID$ init & 65.7 & 80.7 & 85.1 & 88.5 & 17.2 \\
OPD $\to$ GRPO & 68.0 & 82.4 & 86.6 & $\mathbf{89.7}$ & 16.2 \\
$\SFTOOD\to$ GRPO & 70.6 & 82.5 & 86.3 & 88.6 & 15.5 \\
$\SFTID\to$ GRPO & 68.3 & 81.5 & 85.6 & 88.7 & 15.6 \\
\bottomrule
\end{tabular}
\end{table}

The entropy$\to$exploration link, clean in-domain at init, \textbf{breaks OOD}. We searched for a positive, persistent entropy benefit and did not find one. The honest reading: Stage-1 robustly sets the entropy regime; that regime has a real exploration consequence whose scope is the in-domain initialization only --- not the RL endpoint, not OOD.

\subsection{Method or confound?}

A reviewer-critical question (\S2) is whether the entropy gap is a \emph{method} effect or a consequence of OPD seeing the hard tail and the teacher's token-level uncertainty, which the rejection-sampled SFT set excludes. Persistence during RL and recurrence across software stacks show that the observation is not a short-lived logging artifact. The objective-level argument in \S3 supplies a plausible mechanism. Neither observation identifies causality, however, because every OPD run retains the same coverage difference. We therefore treat ``OPD versus SFT'' as an association in this study; the data-aligned $807$-problem control is required for a method-specific claim.

\section{Related Work}

Two-stage SFT$\to$RL pipelines are common in multimodal reasoning, ranging from curriculum training to large-scale open recipes \citep{currreft2025,openmmreasoner2025}. VOLD combines OPD and GRPO with a cross-modal text teacher and finds cold-start alignment important \citep{vold2025}; our setting instead uses a same-modality VLM teacher and sequential stages. Recent text-only analyses separate the effects of state distribution, KL direction, and trajectory source \citep{states2026,decoupling2026}. In particular, reverse-KL OPD can collapse entropy in other settings \citep{decoupling2026}. Our higher-entropy observation is therefore best read as a boundary condition for those results, not a contradiction or a general consequence of teacher modality.

\section{Limitations}

\begin{itemize}\setlength{\itemsep}{2pt}
\item \textbf{Single task, single teacher.} Findings rest on Geometry3K and one 72B same-modality teacher; the narrow endpoint band (F2) is task-specific and the observed entropy ordering (F3) is teacher- and setup-specific. A cross-modal-teacher control is a valuable follow-up.
\item \textbf{Difficulty-coverage confound (primary open issue).} OPD trains on all $2{,}101$ problems; the rejection-sampled SFT set covers only $807$, omitting the $1{,}294$-problem hard tail. \S6.4 argues the entropy gap is not purely a data artifact, but the decisive test --- re-running OPD on the same $807$-problem set --- is not yet done and is the most important follow-up.
\item \textbf{Bounded positive result.} We do not show the entropy regime improves the RL endpoint or OOD; we show it does not, within our tests. The contribution is delineation, not a win.
\item \textbf{Theory scope.} The \S3 discussion is objective-level intuition, not a convergence, causal, or finite-sample guarantee.
\item \textbf{Uncertainty scope.} Problem-level bootstrap intervals quantify benchmark sampling uncertainty, but there is only one decoding seed per checkpoint and no independent training seeds.
\item \textbf{Eval calibration.} A $\approx5$-point internal$\leftrightarrow$offline Geometry3K gap (rollout-vs-offline code paths, not versioning) limits cross-pipeline point comparisons; we avoid it by staying within one pipeline per comparison.
\end{itemize}

\section{Conclusion}

In this small-data VLM study, Stage-1 choice is strongly associated with the entropy regime carried into RL, while in-domain endpoints occupy a narrow band and OOD retention depends heavily on the SFT recipe. OPD has higher entropy and a higher in-domain initialization pass@16 than one SFT baseline, but the smaller SFT contrast is uncertain and no advantage is detected after RL or on MathVista. The practical takeaway is deliberately bounded: entropy is the clearest observed difference between these warm-starts, not evidence that OPD is a universally better RL initialization.

\bibliographystyle{plainnat}
\bibliography{references}

\end{document}